\pdfoutput=1
\documentclass[11pt]{article}
\usepackage[utf8]{inputenc}
\usepackage[letterpaper,margin=1in]{geometry}
\usepackage{times}

\usepackage{amsmath,amssymb,amsthm,mathtools}
\usepackage{graphicx,booktabs,tabularx,array,longtable,multirow}
\usepackage{algorithm,algpseudocode}
\usepackage{listings,xcolor}
\usepackage{caption,subcaption}
\usepackage[numbers,sort&compress]{natbib}
\usepackage[colorlinks=true,allcolors=blue]{hyperref}
\usepackage{authblk}

\usepackage{setspace}
\newtheorem{theorem}{Theorem}

\newcolumntype{Y}{>{\raggedright\arraybackslash}X}

\title{Large Language Models for Structured Clinical Data Analysis:\\ Dual-Agent Grounding and Validation}
\author[1]{Erfan D. Dehkalani\thanks{Corresponding author: \href{mailto:Erfan.Darzi@childrens.harvard.edu}{Erfan.Darzi@childrens.harvard.edu}}}
\author[2]{Seetha Shankaran}
\author[3]{Abbot R. Laptook}
\author[4]{C. Michael Cotten}
\author[1]{P. Ellen Grant}
\author[1]{Yangming Ou}
\affil[1]{Fetal-Neonatal Neuroimaging and Developmental Science Center, Boston Children's Hospital, and Harvard Medical School, Boston, MA, USA}
\affil[2]{Department of Pediatrics, Wayne State University School of Medicine, Detroit, MI, USA}
\affil[3]{Department of Pediatrics, Women \& Infants Hospital of Rhode Island, and Warren Alpert Medical School of Brown University, Providence, RI, USA}
\affil[4]{Division of Neonatology, Department of Pediatrics, Duke University School of Medicine, Durham, NC, USA}
\date{}

\begin{document}
\maketitle

\begin{abstract}
\noindent\textbf{Objective:} To develop and characterize CLEAR-Med, a dual-agent framework for natural-language analysis of structured clinical data that separates SQL-based invocation from independent validation.

\textbf{Methods:} CLEAR-Med uses one agent to translate a question into executable Structured Query Language (SQL), retain the executed query and database result, and produce a draft. Deterministic checks and a separately invoked cross-provider Validation Agent then accept the draft, request one bounded repair, or abstain. We formalized the system as a bounded selective pipeline and evaluated CLEAR-Med's configuration and scalability, and the Invocation Agent's accuracy and consistency on a 25-query development benchmark, using a harmonized 21-site neonatal hypoxic-ischemic encephalopathy table containing 532 de-identified infant records and approximately 1,300 variables.

\textbf{Results:} CLEAR-Med completed all six nominal scalability configurations, including $500\times1300$. Across 25 development-benchmark queries repeated five times, the Invocation Agent answered 83 of 125 responses correctly (66.4\%; query-cluster bootstrap 95\% CI, 48.0--83.2\%), compared with 15 of 125 (12.0\%; 95\% CI, 3.2--22.4\%) for the ungrounded ChatGPT baseline, a paired improvement of 54.4 percentage points (95\% CI, 36.8--72.0\%).

\textbf{Conclusion:} CLEAR-Med provides a general architecture for traceable analysis of structured clinical data: numerical claims remain linked to executed SQL, and unresolved cases can fail closed. The reported experiments characterize CLEAR-Med's configuration and scalability and the Invocation Agent's accuracy, while the formal analysis establishes the encoded-property guarantee of the complete control flow; a prospective full-pipeline evaluation of the validation and abstention stages is the next stage of this work.
\end{abstract}


\section{Introduction}

Large language models (LLMs) can summarize biomedical prose, but governed clinical tables require a different kind of grounding: the answer must follow the schema, cohort definition, denominator, and executed calculation. An ungrounded model can produce a plausible numerical statement without access to the rows that support it, and fluent wording does not expose whether the value came from the intended population \cite{ref1,ref2}.

We use neonatal hypoxic-ischemic encephalopathy (HIE) as the study domain. HIE follows impaired oxygen and blood supply around birth and can be associated with neonatal brain injury visible on magnetic resonance imaging \cite{ref3}. Its trial data combine treatment assignments, physiological measurements, imaging variables, and longitudinal outcomes, making cohort selection and denominator fidelity central to retrospective analysis.

The analysis table harmonizes two multicenter Neonatal Research Network trials across 21 sites and contains 532 de-identified infant records with approximately 1,300 variables per record \cite{ref4,ref5}. Each row represents one participant, and all variables follow NRN definitions.

Three technical problems follow. First, a relational answer depends on exact columns, joins, filters, and aggregation operators. Second, serializing a $532 \times 1300$ clinical table into a prompt discards the database interface and does not scale as a general query mechanism. Third, the system must translate a natural-language question into SQL and translate the executed result back into an answer without changing the cohort or numerical meaning. The executed database must therefore remain the numerical source of record.

We propose \emph{CLEAR-Med} (Clinical Enhanced Analysis and Retrieval System for Medicine), a dual-agent framework that coordinates two separately invoked components with distinct roles. The \emph{Invocation Agent} generates and executes SQL against the structured HIE database and produces a draft response. The \emph{Clinical Validation Agent} checks the draft, its executed SQL, and its database result against deterministic and clinical constraints before the system accepts, repairs, or abstains.

\noindent\textbf{Contributions.} This work makes three contributions:
\begin{enumerate}
  \item[(1)] a provenance-preserving dual-agent architecture that separates SQL generation and execution from deterministic checks, independent validation, bounded repair, and fail-closed abstention (Figure~1 and Algorithm~1);
  \item[(2)] an oracle-scored evaluation of CLEAR-Med's configuration and scalability, and of the Invocation Agent's accuracy and consistency, across 25 clinical query templates forming a development benchmark with repeated trials and task-level breakdowns (Section~3.7); and
  \item[(3)] a bounded selective-decision model and a property-specific encoded-validity guarantee for the complete dual-agent control flow (Section~3.2 and Appendix~A.4).
\end{enumerate}

The formal contribution is deliberately property-specific. If faithful database execution and sound executable checks establish a stated property, CLEAR-Med can emit only an answer that satisfies that property; unresolved cases abstain. This guarantee does not extend to clinical judgments that have not been encoded as executable checks. Section~3.2 states the result, and Appendix~A.4 gives its compact proof.

\begin{table}[htbp]
\centering
\caption{Statement of Significance}
\label{tab:1}
\small
\begin{tabularx}{\textwidth}{@{}>{\raggedright\arraybackslash\bfseries}p{0.2\textwidth}Y@{}}
\toprule
Problem or Issue & Natural-language analysis of governed clinical tables can produce fluent numerical answers without exposing the cohort, calculation, or supporting rows. \\
\midrule
What is Already Known & Language models can generate SQL and use external tools, but query generation alone does not establish that a reported number preserves the executed result and intended clinical constraints. \\
\midrule
What this Paper Adds & CLEAR-Med separates schema-guided SQL invocation from deterministic checks and an independently invoked Validation Agent. The pipeline retains SQL provenance, permits one bounded repair, abstains on unresolved failures, and provides a property-specific encoded-validity guarantee. \\
\midrule
Who would benefit from the new knowledge in this paper & Biomedical-informatics researchers and clinical-data teams building auditable natural-language interfaces to governed relational data. \\
\bottomrule
\end{tabularx}
\end{table}

\section{Related Work}

Biomedical language models have been adapted for medical dialogue, consultation, multilingual question answering, and domain knowledge \cite{ref6,ref7,ref8,ref9,ref10,ref11,ref12}. Healthcare-agent research extends this line toward planning, tool use, and multi-agent collaboration; a recent JBI survey organizes those systems around agent capabilities, applications, and evaluation requirements \cite{ref13}. CLEAR-Med addresses a narrower informatics problem: producing an auditable numerical answer from a governed relational table.

Clinical text-to-SQL systems translate natural-language requests into database operations. UniQA used a unified encoder--decoder architecture for complex electronic health-record questions \cite{ref14}; MedT5SQL studied healthcare-domain text-to-SQL conversion \cite{ref15}; and Criteria2Query 3.0 used a generative language model to construct executable cohort queries from clinical-trial eligibility criteria \cite{ref16}. Other work has examined context-aware pharmacovigilance SQL and generalization across medical text-to-SQL datasets \cite{ref17,ref18}. CLEAR-Med treats query generation as the first stage rather than the endpoint: the executed SQL, raw database result, and natural-language draft form an evidence record for a separate validation decision.

Document-retrieval RAG grounds generation in retrieved passages \cite{ref19,ref20}, while tool-using language models can select and invoke external functions \cite{ref21}. A relational aggregate requires a different evidentiary chain because its value is determined by the selected table, cohort predicate, denominator, and aggregation operator. The dual-agent design combines tool-mediated execution with explicit provenance, executable checks, independent review, bounded repair, and abstention; the formal analysis distinguishes guarantees for encoded properties from residual clinical judgment.

\section{Methods}

\subsection{Overview}

\emph{CLEAR-Med} is a dual-agent framework for natural-language analysis of structured disease-specific clinical data. Figure~1 shows the two-agent workflow. Section~3.3 introduces the SQLite database, Sections~3.4 and 3.5 describe the two agents, and Section~3.6 specifies the complete control flow.

\subsection{Formal Dual-Agent Guarantee}

Let $x$ be a question, $\mathcal{D}$ the fixed database, and $z_j = (q_j, d_j, r_j, \mathbf{c}_j)$ the SQL, executed result, draft, and checker outcomes at attempt $j \in \{0, 1\}$. CLEAR-Med maps each $z_j$ to \textsf{accept}, \textsf{repair}, or \textsf{abstain}, permits at most one repair, and emits $Y \in \{r_0, r_1, \bot\}$, where $\bot$ denotes abstention.

\begin{theorem}[Bounded encoded-validity guarantee]
\label{thm:1}
Let each executable checker $C_k$ be sound for its encoded property $P_k$, so $C_k(z) = 1 \Rightarrow P_k(z) = 1$ for every reachable state. Under faithful database execution and Algorithm~1's fail-closed control flow,
\begin{equation}
\label{eq:1}
N \le 2, \qquad Y \neq \bot \Longrightarrow \bigwedge_{k=1}^{K} P_k(z_\tau), \qquad \Pr_{\mu}\Biggl(Y \neq \bot,\ \neg \bigwedge_{k=1}^{K} P_k(z_\tau)\Biggr) = 0
\end{equation}
for any distribution $\mu$ over admissible questions, where $N$ is the number of SQL-generation attempts and $\tau \in \{0, 1\}$ is the accepted attempt. Every unresolved path terminates in $\bot$.
\end{theorem}

The result follows because acceptance requires every checker to pass, checker soundness transfers those passes to their encoded properties, and the only other terminal outcome is abstention after zero or one repair. Appendix~A.4 gives the proof and the risk--coverage definitions used in evaluation.

\begin{figure}[htbp]
\centering
\includegraphics[width=\textwidth]{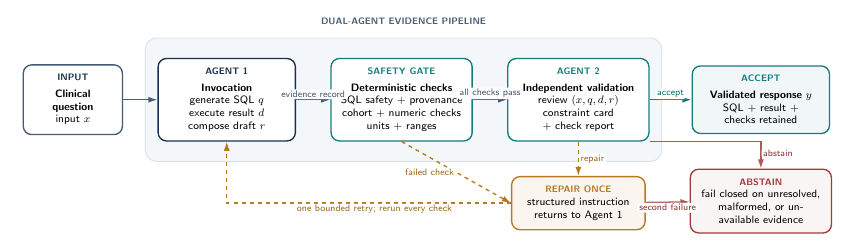}
\caption{CLEAR-Med's complete dual-agent architecture. Agent 1 produces an SQL-grounded evidence record; deterministic checks and Agent 2 independently validate that record before the system accepts with provenance, performs one bounded repair, or fails closed with explicit abstention.}
\label{fig:1}
\end{figure}

\subsection{Convert Excel to SQLite Database}

Our harmonized 21-site HIE database was originally stored as an Excel table with 532 participant rows and approximately 1,300 variable columns. This exact row count describes the analysis table; the $500 \times 1300$ size used later is a nominal synthetic scalability configuration.

We converted the Excel table into an SQLite database---a lightweight, serverless, single-file relational engine---to enable querying via Structured Query Language (SQL).

Each de-identified participant occupies a row and each variable a column; site information is likewise de-identified. All variables follow established Neonatal Research Network (NRN) protocols, so direct transfer to datasets with different variable definitions may require further adaptation and validation.

Beyond more expressive queries than Excel (worked examples appear in Appendix~A), SQLite enforces a schema---table definitions, data types, and inter-variable relationships---which improves data consistency and retrieval. A minimal \texttt{HIEpatient} schema illustrates this:

\begin{lstlisting}[language=SQL,frame=none]
CREATE TABLE HIEpatient (
    patientID INTEGER PRIMARY KEY ,
    birthYear INTEGER ,
    oneMinApgar INTEGER ,
    expertMRIScore TEXT ,
    normalOutcomeOrNot TEXT
);
\end{lstlisting}

\noindent This 5-column schema scales to the thousands of columns in the HIE dictionary and encodes constraints such as the \texttt{PRIMARY KEY} on \texttt{patientID}. Querying it requires translating clinicians' natural-language questions into SQL---the role of the Invocation Agent described next.

\subsection{Invocation Agent}

\begin{figure}[htbp]
\centering
\includegraphics[width=\textwidth]{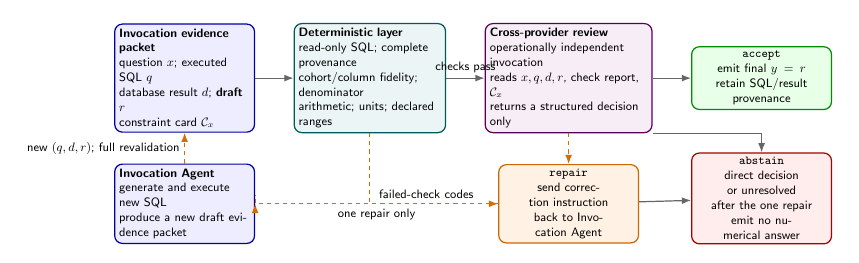}
\caption{Detailed validation and repair flow. The Invocation Agent emits only a draft response $r$ in an evidence packet. The validation stage never edits a numerical answer: a repair returns a structured instruction to SQL generation and execution, after which the entire deterministic and cross-provider validation sequence runs again.}
\label{fig:2}
\end{figure}

The Invocation Agent (Figure~2) takes two inputs---a natural-language query $x$ (e.g., ``What is the association between cord blood pH and neurodevelopmental outcome in HIE patients?'') and the disease-specific database $\mathcal{D}$ with its schema $\mathcal{S}$---and produces a draft response through four stages:
\begin{itemize}
  \item[*] \textbf{Parsing and planning} -- take the natural-language query, extract relevant tables and fields, load the invocation model, and read the schema;
  \item[*] \textbf{SQL generation} -- translate the natural-language query into a schema-aware SQL query $q$;
  \item[*] \textbf{SQL execution against the clinical database} -- execute $q$ against SQLite and retrieve result $d$; and
  \item[*] \textbf{Draft response} -- interpret $d$ as a natural-language draft $r$. This is not the final validated response $y$.
\end{itemize}

At its core, the agent uses LangChain's \texttt{SQLDatabaseToolkit} to generate, execute, and interpret SQL from natural-language inputs, with built-in error-checking for invalid syntax or empty result sets. The design follows \cite{ref21}, in which language models self-supervise external tool use; unlike encoder--decoder EHR question-answering systems that require extensive fine-tuning \cite{ref14}, ours needs none. This mechanism is schema-guided agentic text-to-SQL, not conventional document-retrieval RAG: schema queries place table definitions, column names, and data types in the model context, while the executed relational database remains the source of numerical evidence.

\subsection{Clinical Validation Agent}

The Clinical Validation Agent (Figure~2) converts the Invocation Agent record $(q, d, r)$---the executed SQL $q$, raw result $d$, and draft response $r$---into a final response $y$. Validation first applies deterministic checks for read-only SQL, complete SQL/result provenance, cohort and column fidelity, non-missing denominators, arithmetic, units, and declared ranges. A cross-family language model then independently reviews the question, SQL, result, draft, deterministic-check report, and prospectively authored clinical constraint card. It must return one structured decision: \texttt{accept}, \texttt{repair}, or \texttt{abstain}. Here, \emph{independent} denotes operational separation: the validator is a separately invoked, cross-provider model that did not generate or execute the original SQL.

An accepted response retains the executed SQL provenance. A repair decision cannot directly author a replacement number; instead, it supplies a structured correction instruction to the Invocation Agent, which generates and executes new SQL and undergoes the complete validation sequence again. CLEAR-Med permits exactly one repair. Any second failure, missing evidence, malformed validator response, or unavailable validator produces an explicit abstention. This fail-closed design distinguishes factual support by the database from broader clinical plausibility and prevents a plausible but unsupported number from passing validation.

Theorem~1 formalizes this fail-closed guarantee for encoded properties. The reported empirical analyses characterize CLEAR-Med's configuration, scalability, accuracy, and consistency.

\subsection{Pseudocode for \textup{CLEAR-Med}}

\noindent\textbf{Notation.} Here, $x$ is the question, $q$ the executed SQL, $d$ its database result, $r$ the Invocation Agent's draft, $v$ the deterministic-check report, $y$ the final output, and $\mathcal{D}$ the fixed database. Acceptance sets $y = r$; an unaccepted draft is never emitted.

Algorithm~1 summarizes the provenance-preserving invocation, independent validation, single repair, and fail-closed abstention workflow.

\begin{algorithm}[htbp]
\caption{CLEAR-Med Dual-Agent Workflow}
\label{alg:1}
\begin{algorithmic}[1]
\Require Natural language query $x$; domain schema $\mathcal{S}$; database $\mathcal{D}$; clinical constraints $\mathcal{C}$; invocation model $L_I$; validation model $L_V$
\Ensure Validated clinical response $y$ or explicit abstention
\State Initialize read-only database tools and load the reviewed constraint card $\mathcal{C}_x$
\For{$j \in \{0, 1\}$}
  \State $(q_j, d_j, r_j) \gets \Call{InvocationAgent}{x, \mathcal{S}, \mathcal{D}, L_I}$
  \State $v_j \gets \Call{DeterministicChecks}{x, q_j, d_j, r_j, \mathcal{C}_x}$
  \If{$v_j$ fails}
    \State $a_j \gets \texttt{repair}$ with the failed-check codes
  \Else
    \State $a_j \gets \Call{ClinicalValidationAgent}{x, q_j, d_j, r_j, v_j, \mathcal{C}_x, L_V}$
  \EndIf
  \If{$a_j = \texttt{accept}$}
    \State $y \gets r_j$
    \State \textbf{return} $y$ with $(q_j, d_j, v_j)$ provenance
  \ElsIf{$a_j = \texttt{abstain}$ or $j = 1$}
    \State \textbf{return} explicit abstention with failure codes
  \Else
    \State append the structured repair instruction to the next invocation
  \EndIf
\EndFor
\end{algorithmic}
\end{algorithm}

\setcounter{section}{3}\setcounter{subsection}{6}
\setcounter{equation}{1}
\setcounter{figure}{2}
\setcounter{table}{1}

\subsection{Evaluation Design}
The reported evaluation comprises three analyses: configuration screening, an end-to-end scalability comparison, and an oracle-scored development benchmark of Invocation Agent accuracy and consistency.

\paragraph{Model details.} The configuration screen compared ChatGPT Thinking (OpenAI) with ChatGPT Instant (OpenAI) at temperature 0.7. The dual-agent architecture pairs the selected ChatGPT Thinking invocation configuration with Claude through the Anthropic API at temperature 0 for independent validation using a structured decision schema. The development benchmark measures the Invocation Agent; the validation and abstention stages are specified here and evaluated prospectively in a full-pipeline study. No model training or fine-tuning was performed.

\paragraph{Benchmark set and scoring.} The development benchmark contained 25 natural-language clinical questions covering counts, percentages, group comparisons, odds ratios, and multi-condition filters over the harmonized HIE database. Each question was posed five times to the Invocation Agent and to an ungrounded ChatGPT baseline, yielding 125 responses per condition. Responses were scored against deterministic oracles: percentages and counts within 5\%, percentage-point differences within 2 points, and odds ratios within 10\%. Query-template bootstrap resampling preserved the five repeated trials when estimating confidence intervals and the paired accuracy difference.

\noindent Coverage, accepted-output accuracy, and incorrect-emission risk for the validation and abstention stages are defined in Appendix A.4.3 as the endpoints of that full-pipeline evaluation.

\subsection{Optimization of CLEAR-Med}

\subsubsection{Choosing the ChatGPT configuration}
We compared ChatGPT Thinking (OpenAI) and ChatGPT Instant on the time-based and operational diagnostics defined in Table A.1, Panel A. Where speed and successful SQL generation conflicted, we prioritized successful generation over latency.

\subsubsection{Optimizing other hyperparameters}
We screened four hyperparameters---Max Iterations (10/50/100), Max Execution Time (10/30/60\,s), Agent Type (openai-tools vs.\ tool-calling), and Top K (10/50/100)---varying each in isolation. Table A.1, Panel C, defines the six recorded diagnostics. We first required successful SQL generation and execution, then preferred fewer errors and lower execution cost; SQL length and complexity serve as descriptive query characteristics.

\subsection{End-to-End Scalability Comparison}

\subsubsection{Language-Model Comparators and Interface}
For the end-to-end scalability screen, we selected five language models spanning biomedical, medical-dialogue, and general-purpose configurations. All systems used the same six nominal table sizes, five-minute timeout, and success definition. CLEAR-Med retained each table in SQLite and used ChatGPT Thinking (OpenAI) with schema-guided SQL tools; the comparison models ran on the hosted HPC GPU allocation and processed a serialized table in context. Table 2 compares the tested interfaces end to end.

\begin{itemize}
  \item \textbf{OpenBioLLM-8B} (\texttt{aaditya/Llama3-OpenBioLLM-8B}) \cite{ref6}: an 8B biomedical language model.
  \item \textbf{open-bio-med-8B} (\texttt{timberrific/open-bio-med-merge}) \cite{ref8}: an 8B biomedical language model.
  \item \textbf{Med-ChimeraLlama-3-8B} (\texttt{ChenWeiLi/Med-ChimeraLlama-3-8B\_SHERP}) \cite{ref7}: an 8B medical language model.
  \item \textbf{Daredevil-8B} (\texttt{mlabonne/Daredevil-8B-abliterated}) \cite{ref22}: a general-purpose 8B language model.
  \item \textbf{DISC-MedLLM} \cite{ref11}: a medical-dialogue language model.
\end{itemize}

Exact public identifiers are listed for unambiguous model identification. These five comparators appear in the scalability screen (Section 3.9.2); accuracy and consistency are reported for the separate benchmark in Section 3.9.3.

\subsubsection{Scalability Metric}
The scalability grid contains six row-by-column configurations: $50 \times 50$, $50 \times 100$, $100 \times 50$, $100 \times 100$, $50 \times 1000$, and $500 \times 1300$. A configuration is successful only when it completes within 300 seconds without a GPU out-of-memory failure. We report the percentage of the six configurations completed, the largest successful configuration by cell count, and elapsed time at that configuration.

\subsubsection{Accuracy and Consistency Metrics}
CLEAR-Med's second objective was accurate and consistent numerical analysis on real HIE data. For numeric prediction $x_{q,i}$ and nonzero deterministic oracle $GT_q$, we computed signed relative error $e_{q,i}$, its mean absolute value over defined predictions, and the percentage of all attempts within 5\% of the oracle:
\begin{equation}
\begin{aligned}
e_{q,i} &= 100\,\frac{x_{q,i} - GT_q}{|GT_q|}, &\qquad\qquad
\mathrm{Err} &= \frac{1}{|\mathcal{M}|}\sum_{(q,i)\in\mathcal{M}} |e_{q,i}|,\\
\mathrm{GT}_{5\%} &= 100\,\frac{\#\{|x_{q,i} - GT_q| \le 0.05|GT_q|\}}{\text{all attempts}}. &&
\end{aligned}
\label{eq:err-gt5}
\end{equation}
Oracle-zero, missing, and nonnumeric outputs are excluded from Err because relative error is undefined, but missing and nonnumeric outputs count as failures for $\mathrm{GT}_{5\%}$. Repeated-run consistency is summarized by the range and standard deviation of $e_{q,i}$ within each query, averaged over queries with at least two defined errors.

\section{Results}

\subsection{Optimization of CLEAR-Med}

\subsubsection{Choosing the ChatGPT configuration}
We compared ChatGPT Thinking (OpenAI) and ChatGPT Instant on the SQL-generation task (Table A.1, Panel B). Agent invocation dominated runtime: ChatGPT Instant invoked in 29.23\,s versus 167.44\,s for ChatGPT Thinking (a 5.7$\times$ speedup). ChatGPT Instant produced three errors, hit the iteration limit, and returned no valid SQL, whereas ChatGPT Thinking completed without errors and generated a usable query. Both issued two database invocations, but only ChatGPT Thinking produced a usable result, indicating that optimization should target the invocation step.

Setup, database, and file-loading times were negligible relative to invocation. We therefore selected ChatGPT Thinking (OpenAI) because it completed the SQL-generation task successfully.

\subsubsection{Setting up other key hyperparameters}
\indent\textbf{Max iterations.}\ \ Varying the iteration limit (10/50/100; ChatGPT Thinking (OpenAI), openai-tools, Top K$\,=\,$10), the 50-iteration setting was fastest (134.19\,s), used one database invocation, and retained the highest observed SQL-complexity score (6). The 10-iteration setting was slowest and required three database invocations, while 100 iterations produced a substantially simpler query. Under the stated rule, we selected Max Iterations$\,=\,$50 for the reported configuration.

\medskip
\textbf{Max execution time.}\ \ Varying the time limit (10/30/60\,s), errors remained constant at three per run, but only the 60\,s setting reached a database invocation. We therefore selected 60\,s. Total runtime can exceed the nominal per-step limit because the agent may complete or retry other steps before termination.

\medskip
\noindent\textit{Setting Agent Type.} In the one-factor screen (Table A.1, Panel D), openai-tools completed the task in roughly half the time of tool-calling (80.57\,s vs.\ 165.39\,s); both invoked the database once. We chose openai-tools for its lower execution cost.

\medskip
\noindent\textit{Setting Top K Threshold.} In the one-query screen, Top K$\,=\,$10 was fastest (119.01\,s) and retained more SQL operations than 50 or 100 (complexity 16 vs.\ 2). We therefore retained Top K$\,=\,$10 for the reported configuration. This screen assesses execution and query structure, not answer correctness; numerical validity is evaluated against deterministic oracles and by the complete validation pipeline.

\subsection{End-to-End Scalability Results}

\subsubsection{Scalability}
CLEAR-Med was the only evaluated system to process the largest nominal $500 \times 1300$ test table (Table 2), using $\approx$19.71\,GB of the 39.56\,GB GPU memory; four comparison models reached GPU memory limits beyond $100 \times 100$ ($\approx$16,000 tokens), and one exceeded the timeout beyond $50 \times 50$. Configuring the largest table took 28\,s, comparable to the 2--26\,s required by the comparison models for a table 1/65 its size. CLEAR-Med completed six of six fixed configurations, versus four of six or one of six for the comparison models, under the deployment interfaces described in Section 3.9.

\begin{table}[htbp]
\caption{End-to-end scalability across six fixed table configurations under the deployment interfaces described in Section 3.9. The table reports deterministic completion status and elapsed time for each tested configuration.}
\label{tab:2}
\centering
\footnotesize
\newcommand{\ok}{\textcolor{green!60!black}{$\checkmark$}}
\newcommand{\no}{\textcolor{red}{$\times$}}
\begin{tabular}{lccccccc c}
\toprule
\textbf{System} & 50$\times$50 & 50$\times$100 & 100$\times$50 & 100$\times$100 & 50$\times$1000 & 500$\times$1300 & \begin{tabular}[b]{@{}c@{}}\textbf{Max}\\\textbf{Time (s)}\end{tabular} & \textbf{Notes} \\
\midrule
OpenBioLLM-8B \cite{ref6}          & \ok & \ok & \ok & \ok & \no & \no & 2.19  & CUDA OOM \\
open-bio-med-8B \cite{ref8}        & \ok & \ok & \ok & \ok & \no & \no & 11.78 & CUDA OOM \\
Med-ChimeraLlama-3-8B \cite{ref7}  & \ok & \ok & \ok & \ok & \no & \no & 26.05 & CUDA OOM \\
Daredevil-8B \cite{ref22}          & \ok & \ok & \ok & \ok & \no & \no & 26.00 & CUDA OOM \\
DISC-MedLLM \cite{ref11}           & \ok & \no & \no & \no & \no & \no & 32.86 & Timeout \\
\textbf{CLEAR-Med}                 & \ok & \ok & \ok & \ok & \ok & \ok & \textbf{28.12} & -- \\
\bottomrule
\end{tabular}

\smallskip
{\normalsize \ok: Successful execution; \no: Failed execution}

\medskip
All experiments were conducted on a hosted HPC GPU node with 39.56\,GB of available device memory.\\
CUDA OOM: Memory allocation exceeded available GPU memory.\\
Timeout: Execution exceeded 5 minutes without producing results.
\end{table}

\subsection{CLEAR-Med Development-Benchmark Results}
We evaluated 25 natural-language clinical queries over the harmonized HIE database---counts, percentages, group comparisons, odds ratios, and
multi-condition filters spanning the Optimizing Cooling and Late Hypothermia cohorts---each posed five times ($K = 5$), yielding 125 responses per condition and 250 responses overall. The comparator was the ChatGPT baseline answering directly from an in-context table without database tools. Every response was scored against a deterministic oracle (percent/count within 5\%; percentage-point differences within 2 points; odds ratios within 10\%).

The Invocation Agent answered correctly on 83 of 125 responses (66.4\%; query-cluster bootstrap 95\% CI 48.0--83.2\%) versus 15 of 125 (12.0\%; 95\% CI 3.2--22.4\%) for the ungrounded ChatGPT baseline (OpenAI API), a paired improvement of 54.4 percentage points (95\% CI 36.8--72.0). Task-type values with their denominators appear in Table 4; difficulty-stratified values appear in Table A.1, Panel E. Across the 12 percentage-valued queries (60 responses per condition), the Invocation Agent was within 5\% of the oracle on 39 of 60 responses versus 4 of 60 for the baseline; the corresponding error and repeated-run dispersion measures appear in Table 3.

\begin{table}[htbp]
\caption{Accuracy and consistency on numerical queries across repeated trials ($K = 5$ repetitions, 12 percentage-valued queries). Err = mean absolute percentage error vs.\ the deterministic data oracle; 5\%-GT = share of responses within 5\% of ground truth; SD and Range are computed across repetitions and averaged over queries. Arrows ($\downarrow$/$\uparrow$) indicate the preferred direction.}
\label{tab:3}
\centering
\small
\begin{tabular}{lcccc}
\toprule
\textbf{Condition} & \textbf{Err (\%)}$^{\downarrow}$ & \textbf{5\%-GT (\%)}$^{\uparrow}$ & \textbf{SD (\%)}$^{\downarrow}$ & \textbf{Range (\%)}$^{\downarrow}$ \\
\midrule
ChatGPT baseline (OpenAI API) & 55.74 & 6.67 & 58.80 & 146.58 \\
\textbf{Invocation Agent} & \textbf{33.67} & \textbf{65.00} & \textbf{35.07} & \textbf{78.55} \\
\bottomrule
\end{tabular}
\end{table}

\begin{table}[htbp]
\caption{Per-task accuracy across 25 queries $\times$ 5 repetitions. $n$ is the number of responses per condition in each subgroup. Overall accuracy ($n = 125$ per condition) is reported with a query-cluster bootstrap 95\% confidence interval.}
\label{tab:4}
\centering
\small
\begin{tabular}{lccc}
\toprule
\textbf{Task type} & \textbf{$\boldsymbol{n}$/condition} & \textbf{ChatGPT baseline} & \textbf{Invocation Agent} \\
\midrule
comparison & 30 & 0.0 & 30.0 \\
count & 25 & 36.0 & 100.0 \\
multi-condition & 20 & 5.0 & 50.0 \\
odds ratio & 10 & 20.0 & 100.0 \\
percentage & 40 & 7.5 & 72.5 \\
\midrule
\textbf{Overall (95\% CI)} & \textbf{125} & 12.0 [3.2, 22.4] & \textbf{66.4 [48.0, 83.2]} \\
\bottomrule
\end{tabular}
\end{table}

\section{Discussion}

\subsection{Principal Findings and Interpretation}
The experiments establish two findings. CLEAR-Med completed all six nominal scalability configurations, including $500 \times 1300$, under the tested SQL-tool interface. Across the 25-query development benchmark, the Invocation Agent achieved 66.4\% accuracy against deterministic numerical oracles, compared with 12.0\% for the ungrounded ChatGPT baseline. The comparison reflects the full invocation configuration, combining database access, schema guidance, and prompt design.

CLEAR-Med keeps the relational database as the numerical source of record. Unlike document-retrieval RAG, which retrieves passages for synthesis, the Invocation Agent exposes schema metadata, generates SQL, executes it, and retains the query and result with the draft. The second agent then evaluates that evidence record through deterministic checks and a separately invoked cross-provider review. This division supports governed retrospective analysis and hypothesis generation.

ChatGPT Thinking (OpenAI) was slower than ChatGPT Instant but produced usable SQL in the reported task. The one-factor screens identified the operating settings used for the development configuration.

\subsection{Comparison with Related Approaches}
Clinical text-to-SQL work has shown that language models can translate clinical criteria and questions into executable queries \cite{ref14,ref15,ref16}. CLEAR-Med extends that interface into a selective evidence pipeline: query generation, execution, answer composition, deterministic checking, and independent validation are represented as distinct operations with explicit terminal decisions. The architectural claim is therefore not that a second model substitutes for a database oracle, but that a separately invoked validator can inspect the complete evidence record after deterministic properties have been checked and can trigger bounded repair or abstention.

This distinction also separates CLEAR-Med from passage-based retrieval augmentation. A retrieved passage remains evidence that must be interpreted, whereas a relational aggregate is determined by executable operations over a fixed database. Retaining the SQL, result, check report, and validation decision makes the provenance chain inspectable and provides a natural unit for auditing structured-data analyses.

\subsection{Applicability}
The demonstrated use case is retrospective, research-oriented analysis of a harmonized neonatal HIE schema. The same architecture can be configured for other governed relational clinical datasets by replacing the schema description, executable constraint cards, and deterministic oracles while retaining the invocation--validation control flow. Such transfer calls for dataset-specific validation.

\subsection{Limitations}
The study remains limited to one harmonized HIE schema, sequential rather than joint configuration screening, and a 25-query development benchmark whose errors informed prompt and oracle corrections. Finally, the conditional guarantee covers only properties represented by sound executable checkers under faithful database execution; open-ended clinical reasoning, individualized prognosis, and transfer to other schemas remain outside the demonstrated scope.

\section{Conclusion}
CLEAR-Med provides a dual-agent, schema-guided framework for structured clinical data analysis with SQL provenance, deterministic checks, independent validation, bounded repair, and explicit abstention. The component studies show that CLEAR-Med's SQL interface can scale to the full nominal HIE table and that its Invocation Agent substantially outperformed an ungrounded ChatGPT baseline on oracle-scored development queries. Held-out testing and a full-pipeline evaluation of the validation and abstention stages form the next stage of this work. By making acceptance contingent on an inspectable evidence record, the architecture defines a general route from natural-language questions to auditable numerical answers in governed relational data.

\appendix
\renewcommand{\thetable}{A.\arabic{table}}\setcounter{table}{0}
\renewcommand{\theequation}{A.\arabic{equation}}\setcounter{equation}{0}
\newtheorem{example}{Example}
\renewcommand{\theexample}{A.\arabic{example}}

\section{Extended Methods and Supporting Material}

\subsection{System and Dataset Summary}

\begin{itemize}
  \item \textbf{Intended use:} research and educational analysis of structured neonatal HIE trial data via natural-language querying.
  \item \textbf{Models:} ChatGPT Thinking (OpenAI) at temperature 0.7 for invocation and Claude through the Anthropic API at temperature 0 for independent validation; no fine-tuning. Requested and API-returned identifiers and access timestamps are retained per run.
  \item \textbf{Data:} de-identified, harmonized NRN HIE database (Optimizing Cooling and Late Hypothermia trials, 21 sites); variables defined by established NRN protocols; access restricted (see Data Availability).
  \item \textbf{Reported metrics:} configuration diagnostics, scalability success, numerical accuracy, oracle error, repeated-run dispersion, and task- and difficulty-stratified accuracy.
  \item \textbf{Safety controls:} every accepted number retains SQL provenance and passes deterministic and independent validation; unresolved or malformed cases abstain after at most one repair.
\end{itemize}

\subsection{Supporting Tables}

\begin{table}[htbp]
\centering
\caption{Supporting definitions and results. Panel A: operational definitions for the efficiency and execution diagnostics used to compare ChatGPT configurations. The final column gives the preferred direction; an em dash denotes a descriptive metric. All time measures are in seconds.}
\label{tab:A1}
\begin{tabularx}{\linewidth}{p{0.29\linewidth}Yc}
\toprule
\textbf{Metric} & \textbf{Operational definition} & \textbf{Goal}\\
\midrule
\multicolumn{3}{l}{\textbf{Time-based diagnostics}}\\
Setup environment time & Time to initialize the computational environment & $\downarrow$\\
Load files to database & Time to load data files into the database & $\downarrow$\\
Setup database time & Time to configure the database for query processing & $\downarrow$\\
Setup language model time & Time to initialize the language model & $\downarrow$\\
Create agent time & Time to instantiate the SQL agent & $\downarrow$\\
Invoke agent time & Time for the agent to process a request and generate SQL & $\downarrow$\\
Execution time & Total duration from invocation start to final response & $\downarrow$\\
Total time & Aggregate duration of all setup and processing steps & $\downarrow$\\
\midrule
\multicolumn{3}{l}{\textbf{Operational diagnostics}}\\
Number of errors & Count of SQL syntax errors and execution failures & $\downarrow$\\
SQL query generated & Binary indicator (yes/no) of successful SQL generation & $\uparrow$\\
Number of database invocations & Count of successful database connections and queries & ---\\
\bottomrule
\end{tabularx}
\end{table}

\begin{table}[htbp]
\centering
\addtocounter{table}{-1}
\caption{Supporting definitions and results (continued). Panel B: efficiency and SQL-agent diagnostics for the two ChatGPT configurations. Arrows ($\downarrow$/$\uparrow$) indicate the preferred direction.}
\small
\textit{Function execution times (seconds)}\\[2pt]
\begin{tabular}{@{}lcccccc@{}}
\toprule
\textbf{Configuration} & \begin{tabular}[b]{@{}c@{}}\textbf{Setup}$^{\downarrow}$\\\textbf{environment}\end{tabular} & \begin{tabular}[b]{@{}c@{}}\textbf{Load files}$^{\downarrow}$\\\textbf{to DB}\end{tabular} & \begin{tabular}[b]{@{}c@{}}\textbf{Setup}$^{\downarrow}$\\\textbf{database}\end{tabular} & \begin{tabular}[b]{@{}c@{}}\textbf{Setup}$^{\downarrow}$\\\textbf{LLM}\end{tabular} & \begin{tabular}[b]{@{}c@{}}\textbf{Create}$^{\downarrow}$\\\textbf{agent}\end{tabular} & \begin{tabular}[b]{@{}c@{}}\textbf{Invoke}$^{\downarrow}$\\\textbf{agent}\end{tabular}\\
\midrule
\begin{tabular}[c]{@{}l@{}}ChatGPT Thinking\\(OpenAI)\end{tabular} & $1.10\times10^{-5}$ & 0.02 & 0.14 & 0.04 & 0.24 & 167.44\\
ChatGPT Instant & $1.10\times10^{-5}$ & 0.02 & 0.14 & 0.05 & 0.004 & 29.23\\
\bottomrule
\end{tabular}

\vspace{6pt}
\textit{SQL-agent task performance}\\[2pt]
\begin{tabular}{@{}lccccc@{}}
\toprule
\textbf{Configuration} & \begin{tabular}[b]{@{}c@{}}\textbf{Execution}$^{\downarrow}$\\\textbf{time (s)}\end{tabular} & \begin{tabular}[b]{@{}c@{}}\textbf{Total}$^{\downarrow}$\\\textbf{time (s)}\end{tabular} & \begin{tabular}[b]{@{}c@{}}\textbf{Number of}$^{\downarrow}$\\\textbf{errors}\end{tabular} & \begin{tabular}[b]{@{}c@{}}\textbf{SQL query}$^{\uparrow}$\\\textbf{generated}\end{tabular} & \begin{tabular}[b]{@{}c@{}}\textbf{Number of}\\\textbf{DB invocations}\end{tabular}\\
\midrule
\begin{tabular}[c]{@{}l@{}}ChatGPT Thinking\\(OpenAI)\end{tabular} & 170.64 & 171.14 & 0 & Yes & 2\\
ChatGPT Instant & 87.00 & 87.23 & 3 & No & 2\\
\bottomrule
\end{tabular}
\end{table}

\begin{table}[htbp]
\centering
\addtocounter{table}{-1}
\caption{Supporting definitions and results (continued). Panel C: diagnostics used in the one-factor-at-a-time configuration screen. The final column gives the preferred direction where applicable; an em dash denotes a descriptive diagnostic.}
\begin{tabularx}{\linewidth}{p{0.29\linewidth}Yc}
\toprule
\textbf{Metric} & \textbf{Operational definition} & \textbf{Goal}\\
\midrule
\multicolumn{3}{l}{\textbf{Performance diagnostics}}\\
Execution time & Duration from invocation start to final response & $\downarrow$\\
Total time & Execution time plus all setup operations & $\downarrow$\\
Number of database invocations & Count of successful database connections and queries & ---\\
Number of errors & Count of SQL syntax errors and execution failures & $\downarrow$\\
\midrule
\multicolumn{3}{l}{\textbf{Query diagnostics}}\\
SQL length & Character count of the generated SQL statement & ---\\
SQL complexity & Sum of joins (J), conditions (C), aggregations (A), and subqueries (S): J+C+A+S & ---\\
\bottomrule
\end{tabularx}
\end{table}

\begin{table}[htbp]
\centering
\addtocounter{table}{-1}
\caption{Supporting definitions and results (continued). Panel D: one-factor-at-a-time configuration screen for CLEAR-Med. The check mark identifies the setting selected for the reported configuration. Selection first required successful SQL generation and execution, then used errors and execution cost; SQL length and complexity are descriptive. An em dash indicates that the diagnostic was not recorded for that factor.}
\scriptsize
\begin{tabular}{@{}llrrrrrrc@{}}
\toprule
\textbf{Factor} & \textbf{Setting} & \textbf{Exec. (s)} & \textbf{Total (s)} & \textbf{Errors} & \textbf{DB inv.} & \textbf{SQL len.} & \textbf{SQL compl.} & \textbf{Selected}\\
\midrule
\multirow{3}{*}{Max iterations} & 10 & 317.12 & 317.30 & -- & 3 & 503 & 5 & \\
 & 50 & 134.19 & 134.71 & -- & 1 & 555 & 6 & \textcolor{green!60!black}{$\checkmark$}\\
 & 100 & 139.08 & 139.28 & -- & 1 & 208 & 2 & \\
\midrule
\multirow{3}{*}{Max execution time} & 10\,s & 38.52 & 38.71 & 3 & 0 & -- & -- & \\
 & 30\,s & 49.90 & 50.13 & 3 & 0 & -- & -- & \\
 & 60\,s & 85.40 & 85.60 & 3 & 1 & -- & -- & \textcolor{green!60!black}{$\checkmark$}\\
\midrule
\multirow{2}{*}{Agent type} & openai-tools & 80.57 & 81.06 & -- & 1 & 208 & 2 & \textcolor{green!60!black}{$\checkmark$}\\
 & tool-calling & 165.39 & 165.64 & -- & 1 & 350 & 3 & \\
\midrule
\multirow{3}{*}{Top K} & 10 & 119.01 & -- & -- & -- & 717 & 16 & \textcolor{green!60!black}{$\checkmark$}\\
 & 50 & 133.16 & -- & -- & -- & 294 & 2 & \\
 & 100 & 134.29 & -- & -- & -- & 196 & 2 & \\
\bottomrule
\end{tabular}
\end{table}

\begin{table}[htbp]
\centering
\addtocounter{table}{-1}
\caption{Supporting definitions and results (continued). Panel E: accuracy by query difficulty. $n$ is the number of responses per condition; the clustered interval for the overall result appears in Table 4.}
\begin{tabular}{@{}lccc@{}}
\toprule
\textbf{Difficulty} & \textbf{$n$/condition} & \textbf{ChatGPT baseline} & \textbf{Invocation Agent}\\
\midrule
easy & 45 & 24.4 & 86.7\\
medium & 25 & 8.0 & 80.0\\
hard & 55 & 3.6 & 43.6\\
\bottomrule
\end{tabular}
\end{table}

\subsection{Illustrative SQL Examples}

The following worked examples illustrate the querying advantages of the SQLite formulation described in Section 3.3.

\begin{example}
Compared to Excel, one advantage of SQLite is that it enables more powerful and flexible querying capabilities, particularly for complex data operations. For instance, in Excel, performing a query to analyze the correlation between neonatal brain injury severity and long-term neurodevelopmental outcomes would require complex formulas or Visual Basic for Applications (VBA) scripting. In contrast, SQLite allows for such operations using a simple SQL query. An example SQL query in SQLite might look like this:
\begin{lstlisting}[language=SQL,frame=none]
SELECT severity,
    AVG(cognitive_score) as avg_cognitive,
    AVG(language_score) as avg_language,
    AVG(motor_score) as avg_motor
    FROM hie_patients
    GROUP BY severity
    ORDER BY severity;
\end{lstlisting}
This query efficiently analyzes the relationship between HIE severity and various neurodevelopmental outcomes, a task that would be cumbersome in Excel.
\end{example}

\begin{example}
Another advantage of SQLite over Excel is its ability to handle large datasets more efficiently. While Excel has limitations on the number of rows or columns it can process, SQLite can manage millions of records with ease. For example, an SQL query to analyze temporal changes in various clinical variables for HIE patients could be written in SQLite as:
\begin{lstlisting}[language=SQL,frame=none]
SELECT patient_id,
    AVG(CASE WHEN time_point <= 6 THEN core_temp END) as avg_temp_0_6h,
    AVG(CASE WHEN time_point > 6 AND time_point <= 24 THEN core_temp END) as avg_temp_6_24h,
    AVG(CASE WHEN time_point > 24 AND time_point <= 48 THEN core_temp END) as avg_temp_24_48h,
    AVG(CASE WHEN time_point > 48 AND time_point <= 72 THEN core_temp END) as avg_temp_48_72h,
    MAX(sarnat_score) as max_sarnat_score,
    MIN(umbilical_cord_ph) as min_umbilical_cord_ph,
    SUM(CASE WHEN treatment = 'cooling' THEN 1 ELSE 0 END) as cooling_sessions,
    AVG(cooling_blanket_temp) as avg_cooling_blanket_temp
    FROM hie_patient_data
    GROUP BY patient_id;
\end{lstlisting}
This query efficiently analyzes and summarizes multiple clinical variables across different time points for HIE patients, including core temperature changes, neurological assessments (Sarnat score), biochemical markers (umbilical cord pH), treatment interventions (cooling sessions), and physiological measurements (cooling blanket temperature). Such a comprehensive analysis would be challenging and potentially crash Excel with a sufficiently large dataset, while SQLite can handle it with ease.
\end{example}

\subsection{Formal Analysis of the Dual-Agent Selective Pipeline}

This appendix supplies the transition semantics and proof underlying Theorem 1, then defines the selective-evaluation quantities for prospective full-pipeline evaluation. It also separates gatewise residual risk without assuming independence between agents and states the predictive-uncertainty identity that motivates deterministic grounding. The guarantee itself applies only to properties represented by sound executable checkers.

\subsubsection{State and Transition Semantics}

For attempt $j\in\{0,1\}$, let $X$ be the question, $Q_j$ the generated SQL, $D_j=\mathrm{Exec}(Q_j,\mathcal{D})$ its faithfully executed result, $R_j$ the draft, and $H_j$ the repair history. The reachable state and conjunction of deterministic checker outcomes are
\begin{equation}
S_j=(j,X,\mathcal{S},Q_j,D_j,R_j,\mathcal{C}_X,H_j),\qquad G_j=\prod_{k=1}^{K}C_k(S_j).
\end{equation}
When $G_j=1$, write the separately invoked validator's decision as $A_j$; its possible actions are \texttt{accept}, \texttt{repair}, and \texttt{abstain}. The fail-closed transition is
\begin{equation}
\Delta_j=
\begin{cases}
\textsf{accept}, & G_j=1,\ A_j=\textsf{accept},\\
\textsf{repair}, & j=0,\ G_j=0 \text{ or } A_j=\textsf{repair},\\
\textsf{abstain}, & \text{otherwise.}
\end{cases}
\end{equation}
Operational independence means that the validator is a separate cross-provider invocation that did not generate $Q_j$ or execute it. It does not imply probabilistic independence between generation and validation.

\subsubsection{Proof of the Encoded-Validity Guarantee}

Let $\mathsf{R}$ denote first-attempt repair. The four terminal paths are first-attempt acceptance, first-attempt abstention, repair followed by acceptance, and repair followed by abstention. They are disjoint and exhaustive, so
\begin{equation}
N=1+\mathbf{1}_{\mathsf{R}}\le 2,\qquad
Y=\begin{cases}
r_\tau, & \text{on acceptance},\\
\bot, & \text{otherwise},
\end{cases}
\qquad \tau\in\{0,1\}.
\end{equation}
If $Y\neq\bot$, acceptance gives $G_\tau=1$, hence $C_k(S_\tau)=1$ for every $k$. Soundness gives $C_k(S_\tau)=1\Rightarrow P_k(S_\tau)=1$, proving the conjunction in Equation (1). The encoded-violation event is therefore empty for every admissible question; integrating its zero indicator under any input distribution $\mu$ proves the probability statement.

\subsubsection{Selective Evaluation Quantities}

For condition $m$, let $E^{(m)}=\mathbf{1}\{Y^{(m)}\neq\bot\}$ indicate emission and let $Z^{(m)}$ indicate oracle correctness. Coverage $C_m$, incorrect-emission risk $R_m$, and accepted-output accuracy $A_m$ are
\begin{equation}
\begin{aligned}
C_m&=\mathbb{E}[E^{(m)}], & R_m&=\mathbb{E}[E^{(m)}(1-Z^{(m)})],\\
A_m&=\Pr(Z^{(m)}=1\mid E^{(m)}=1), & R_m&=C_m(1-A_m).
\end{aligned}
\end{equation}
For $n$ question--repetition trials, the corresponding plug-in estimators are
\begin{equation}
\begin{aligned}
\widehat{C}_m&=\frac{1}{n}\sum_{\ell=1}^{n}E_\ell^{(m)}, & \widehat{R}_m&=\frac{1}{n}\sum_{\ell=1}^{n}E_\ell^{(m)}(1-Z_\ell^{(m)}),\\
\widehat{A}_m&=\frac{\sum_\ell E_\ell^{(m)}Z_\ell^{(m)}}{\sum_\ell E_\ell^{(m)}}, & \widehat{O}_m&=\frac{1}{n}\sum_{\ell=1}^{n}E_\ell^{(m)}Z_\ell^{(m)}.
\end{aligned}
\end{equation}
For conditions $a$ and $b$, $\widehat{\Delta}_R=\widehat{R}_a-\widehat{R}_b$ measures the change in incorrect emission and $\widehat{\Delta}_C=\widehat{C}_b-\widehat{C}_a$ records the accompanying coverage change. Paired bootstrap resampling preserves repeated trials from the same question template. These quantities are the prespecified endpoints for the full-pipeline evaluation.

\subsubsection{Dependence-Aware Residual Risk}

For attempt $j$, let $\mathsf{I}_j$ denote an oracle-incorrect draft, $\mathsf{G}_j$ a deterministic-gate pass, and $\mathsf{L}_j$ validator acceptance, conditional on reaching that attempt ($\mathcal{H}_j$). The residual incorrect-acceptance probability factors exactly as
\begin{equation}
\Pr(\mathsf{I}_j\cap\mathsf{G}_j\cap\mathsf{L}_j\mid\mathcal{H}_j)=\Pr(\mathsf{I}_j\mid\mathcal{H}_j)\,\Pr(\mathsf{G}_j\mid\mathsf{I}_j,\mathcal{H}_j)\,\Pr(\mathsf{L}_j\mid\mathsf{I}_j,\mathsf{G}_j,\mathcal{H}_j).
\end{equation}
This is the conditional chain rule, not an independence assumption. With $\mathsf{B}_j=\mathsf{I}_j\cap\mathsf{G}_j\cap\mathsf{L}_j$, $\mathsf{R}$ denoting first-attempt repair, and $\rho=\Pr(\mathsf{R})$, the two disjoint acceptance branches give
\begin{equation}
\Pr(\mathsf{B})=\Pr(\mathsf{B}_0)+\rho\,\Pr(\mathsf{B}_1\mid\mathsf{R}).
\end{equation}
If the three conditional factors at attempt $j$ are bounded by $\varepsilon_{I,j}$, $\varepsilon_{G,j}$, and $\varepsilon_{L,j}$, respectively, then
\begin{equation}
\Pr(\mathsf{B})\le\varepsilon_{I,0}\varepsilon_{G,0}\varepsilon_{L,0}+\rho\,\varepsilon_{I,1}\varepsilon_{G,1}\varepsilon_{L,1}.
\end{equation}
For an error that violates a property covered by a sound checker, the corresponding deterministic-pass factor is zero. No such conclusion follows for an unencoded clinical judgment, so the bound does not multiply unconditional agent error rates.

\subsubsection{Conditional Predictive Uncertainty}

Let $\Theta$ represent uncertainty over sampled model behavior, $Q$ the generated SQL, $D=\mathrm{Exec}(Q,\mathcal{D})$, and $Y$ the response. For discrete $Y$ with finite conditional entropy,
\begin{equation}
H(Y\mid X,D)=I(Y;\Theta\mid X,D)+H(Y\mid X,D,\Theta).
\end{equation}
The identity follows directly from the definition of conditional mutual information. For a deterministic aggregate, the target is fixed once the database, cohort, and operator are fixed; repeated runs therefore characterize model-induced variability in $(Q,Y)$, whereas genuinely predictive clinical targets may retain irreducible outcome uncertainty. The dual-agent control flow does not eliminate that distinction: it makes executable properties checkable and routes unresolved cases to abstention.

\section*{Acknowledgments}

We gratefully acknowledge the Neonatal Research Network (NRN) within the Eunice Kennedy Shriver National Institute of Child Health and Human Development (NICHD) Neonatal Research Network for providing us access to the Optimizing Cooling (OC) and Late Hypothermia (LH) trial datasets, through RTI International for their role as the Data Coordinating Center in managing these valuable neonatal medical datasets (Scott A. McDonald, Jeanette O. Auman). Data from the 21 sites in these 2 trials were recently harmonized into a unified database at Boston Children's Hospital (Chuan-Heng Hsiao, Anna N. Foster, Rutvi Vyas, Ankush Kesri, Rina Bao, Matheus D. Soldatelli, Aseelah Ashraf, Lena Tran, Janet S. Soul), which is the data analysis center in the Consortium Of MRI Biomarker In Neonatal Encephalopathy (COMBINE). This research was funded, in part, by NIH R61 NS126792 and R21 NS121735.

\section*{Funding}

This work was supported, in part, by the National Institutes of Health [grant numbers R61 NS126792 and R21 NS121735]. The funders had no role in study design; data collection, analysis, or interpretation; manuscript preparation; or the decision to submit the article for publication.

\section*{CRediT authorship contribution statement}

\textbf{Erfan Darzidehkalani:} Conceptualization, Methodology, Software, Formal analysis, Writing -- original draft. \textbf{Seetha Shankaran:} Validation, Writing -- review \& editing. \textbf{Abbot R. Laptook:} Validation, Writing -- review \& editing. \textbf{C. Michael Cotten:} Validation, Writing -- review \& editing. \textbf{P. Ellen Grant:} Supervision, Funding acquisition, Writing -- review \& editing. \textbf{Yangming Ou:} Supervision, Funding acquisition, Writing -- review \& editing.

\section*{Declaration of competing interest}

The authors declare that they have no known competing financial interests or personal relationships that could have appeared to influence the work reported in this paper.

\section*{Ethics approval}

This study is a secondary analysis of data harmonized from two completed multicenter randomized controlled trials conducted by the NICHD Neonatal Research Network: the Optimizing Cooling (OC) trial \cite{ref5} and the Late Hypothermia (LH) trial \cite{ref4}. The original trials were approved by the institutional review boards of participating sites, and informed consent was obtained from a parent or legal guardian of each enrolled infant. For the present study, investigators accessed only fully de-identified records under applicable data-governance arrangements; no new recruitment, intervention, participant contact, or attempt at re-identification occurred. No separate study-specific ethics reference number applies to the present analysis.

\section*{Consent to participate}

Informed consent was obtained from a parent or legal guardian of each neonatal participant in the original OC and LH trials. The present analysis involved no new participant enrollment or contact.

\section*{Consent for publication}

Not applicable. This manuscript does not contain identifiable individual data.

\section*{Sex and gender reporting}

This study evaluates a software pipeline rather than clinical associations or treatment effects. Sex and gender were not used as predictors, outcomes, eligibility criteria, or subgroup variables in the reported benchmark; accordingly, no sex- or gender-stratified analysis was performed.

\section*{Data availability}

The harmonized analytical table used in this study is not publicly redistributed. De-identified source datasets for the Optimizing Cooling and Late Hypothermia trials are available by application through the NICHD Data and Specimen Hub (DASH). Requests are processed through the DASH data-request and data-use-agreement workflow. Access to the study-specific harmonized table remains subject to applicable NICHD Neonatal Research Network and institutional governance.

\bibliographystyle{unsrtnat}
\bibliography{refs}
\end{document}